\documentclass[10pt,twocolumn,letterpaper]{article}

\usepackage{iccv}              

\definecolor{iccvblue}{rgb}{0.21,0.49,0.74}
\usepackage[pagebackref,breaklinks,colorlinks,allcolors=iccvblue]{hyperref}
\usepackage{multirow}
\usepackage{bm}
\usepackage{booktabs}
\def\paperID{7962} 
\def\confName{ICCV}
\def\confYear{2025}

\title{FED-PsyAU: Privacy-Preserving Micro-Expression Recognition via Psychological AU Coordination and Dynamic Facial Motion Modeling}

\author{Jingting Li$^{\text{\textdagger}1,2}$, Yu Qian$^{\text{\textdagger}1,3}$, Lin Zhao$^{1,2}$, Su-Jing Wang*$^{1,2}$\\
$^1$State Key Laboratory of Cognitive Science and Mental Health, Institute of Psychology, \\Chinese Academy of Sciences, Beijing, 100101, China \\
$^2$Department of Psychology, University of the Chinese Academy of Sciences, Beijing, 100049, China\\
$^3$School of Computer, Jiangsu University of Science and Technology, Zhenjiang, 212100, China \\
{\tt\small $^{\text{\textdagger}}$Equal contribution, $^*$Corresponding author (wangsujing@psych.ac.cn)}
}

\begin{document}
\maketitle
\begin{abstract}
Micro-expressions (MEs) are brief, low-intensity, often localized facial expressions. They could reveal genuine emotions individuals may attempt to conceal, valuable in contexts like criminal interrogation and psychological counseling. However, ME recognition (MER) faces challenges, such as small sample sizes and subtle features, which hinder efficient modeling. Additionally, real-world applications encounter ME data privacy issues, leaving the task of enhancing recognition across settings under privacy constraints largely unexplored.
To address these issues, we propose a FED-PsyAU research framework. We begin with a psychological study on the coordination of upper and lower facial action units (AUs) to provide structured prior knowledge of facial muscle dynamics. We then develop a DPK-GAT network that combines these psychological priors with statistical AU patterns, enabling hierarchical learning of facial motion features from regional to global levels, effectively enhancing MER performance. Additionally, our federated learning (FL) framework advances MER capabilities across multiple clients without data sharing, preserving privacy and alleviating the limited-sample issue for each client. 
Extensive experiments on commonly-used ME databases demonstrate the effectiveness of our approach. Our implementation is publicly available at~\href{https://github.com/MELABIPCAS/FED-PsyAU.git}{https://github.com/MELABIPCAS/FED-PsyAU.git}.
\end{abstract}

\section{Introduction}
Micro-expressions (MEs)~\cite{66}, resulting from conflicts between voluntary and involuntary expressions, reveal true emotions, making them valuable in non-contact emotion detection applications like criminal interrogation and psychological counseling.
MEs are characterized by their brief duration, low intensity, localized occurrence, and occasional asymmetry~\cite{ekman1978facial}. These traits not only restrict large-scale data collection and annotation but also make it challenging for deep learning networks to effectively model their motion characteristics.
Some methods focus on regions of interest (ROIs)~\cite{35,LBP-TOP,10,67}, while others utilize attention mechanisms~\cite{68,69,70} and transformer-based architectures~\cite{34,71,72} to capture salient features. Meantime, MEs are often complex manifestations involving the combination of multiple AUs. In contrast, individual action units (AUs) correspond to specific, anatomically-based facial muscle movements.
Focusing on AUs allows us to model more fundamental and consistent facial appearance changes.
Certain approaches use AU labels to construct adjacency matrices for graph convolutional networks (GCNs)~\cite{34,74}. Yet, to the best of our knowledge, these methods often apply AU annotations directly without fully exploiting the anatomical structure and coordinated AU dynamics, limiting the network’s understanding on ME.
\par
To address these challenges, we leverage the theory that distinct neural pathways control upper and lower facial movements, resulting in two different motion patterns. Based on this, we conduct a psychological study to investigate upper/lower facial AU coordination, providing crucial prior knowledge on structural motion relationships underpinning MEs.  Using both psychological priors and data-driven AU patterns, we design a Graph Attention Network (GAT)~\cite{62} based on dynamic prior knowledge, applied separately to the upper and lower regions. A final global GAT captures the topological relationships of muscle movements across the face. The resulting feature, combined with full-face optical flow (OF), is input into a subsequent dual-stream network, effectively integrating structural and dynamic information to enhance ME recognition (MER).

\par
Besides, as mentioned earlier, although MEs have valuable applications in many privacy-sensitive contexts, the sensitivity of biometric data makes it impractical to aggregate ME data across clients for training. Meanwhile, the sample size in a single client is often insufficient to train a robust model. Yet, the challenge of enhancing MER performance across diverse clients while preserving privacy remains largely unexplored.
To address this, we adopt a federated learning (FL) framework, improving performance by merging and updating model parameters across clients without exchanging local data. This approach preserves data privacy and supports the MER practical deployment.
\par
Overall, our work leverages psychological and data-driven priors to enhance the network’s learning of structured ME motion patterns, particularly in limited-sample conditions. Additionally, we introduce FL to advance MER in real-world scenarios. The contributions of this paper are:

\begin{itemize}
\item We conduct a psychological experiment to derive AU coordination between the upper and lower facial regions, providing structural priors for ME and facial expression analysis in computer vision.
\item We design a local-to-global GAT network that integrates psychological priors and data-driven AU patterns in the upper and lower facial regions, improving the network’s ability to learn both local and global structural facial dynamics. This feature, combined with global OF feature in a dual-stream network, enhances MER performance.
\item We employ FL in MER to improve data privacy and overcome small sample size challenges by aggregating features from distributed data-protected sources.
\end{itemize}


\section{Related Works}
\subsection{Micro-expression Recognition}
In recent years, MER has gained increasing attention from computer vision researchers. Traditional machine learning methods primarily rely on handcrafted feature extraction. For instance, Local Binary Pattern (LBP)~\cite{LBP}, effectively utilizes local texture features, has inspired various improvements~\cite{LBP-TOP, 5}. Furthermore, OF~\cite{7} has been explored for capturing object motion. Liu et al.~\cite{8} developed the Main Directional Mean OF (MDMO) method by aligning faces in the OF domain, while Liong et al.~\cite{9} added optical strain features for MER. However, traditional approaches depend heavily on manually crafted features, limiting both accuracy and generalizability.
Meanwhile, deep learning methods, with robust feature extraction capabilities, can automatically learn multi-level representations to capture subtle ME changes. The availability of several spontaneous ME databases, such as CASME II~\cite{12},  SAMM~\cite{15}, CAS(ME)$^3$~\cite{li2022cas3}, DFME~\cite{18}and others~\cite{11,13,ben2021video,17,19}, has accelerated deep learning advancements in MER. Inspired by Liong et al.'s work~\cite{22} on sequence redundancy, many studies~\cite{67,71,88} now use keyframes (onset and apex) as input, reducing model parameters while achieving excellent results.
\par
Despite advancements in preprocessing and feature extraction, deep learning alone has yet to achieve high MER accuracy. As MEs stem from physiological muscle activity, incorporating prior knowledge offers meaningful context. FACS~\cite{ekman1978facial}, which defines expressions through AUs, provides a basic theoretical framework for MER.
Recent studies explore AU-ME relationships to enhance MER performance. Lei et al.~\cite{34} applied GCN~\cite{35} to model AU co-occurrences, especially around the eyebrow and mouth regions. Xie et al.~\cite{37} introduced an AU-assisted GCN, with modules for AU feature extraction and ME image generation. Wang et al.~\cite{wang2024micro} used transformers to learn dynamic AU adjacency, boosting model generalization.
These networks typically rely on AU labels from databases, without fully integrating structured facial muscle movements, AU interrelations, or their links to expressions. In this work, we refine structured ME feature extraction using a local-to-global GAT network, which integrates priors on facial AU coordination from psychological experiments with data-driven movement patterns.

\subsection{Federated Learning}
To assess MER across databases, ME Grand Challenge (MEGC) 2018~\cite{38} introduced the Holdout-database Evaluation and Composite Database Evaluation tasks, utilizing the CASME II and SAMM databases. MEGC 2019~\cite{23} expanded the CDE task to include the SMIC databases. While research on MEs predominantly utilizes these public datasets, existing methodologies do not address the privacy concerns that arise in practical applications. 
\par
FL~\cite{41} could addresses ME data scarcity and privacy concerns through collaborative training. It includes two main phases: local training and global aggregation. Many existing FL works for expressions, while insightful, are often not open-source or use highly customized frameworks (e.g., non-random client data allocation~\cite{qi2022flfve}, specialized parameter aggregation~\cite{salman2022pflfe}) making standard methods more suitable for implementation and comparisons in our study. For instance, McMahan et al.~\cite{41} proposed FedAvg, a basic aggregation method that uses weighted averaging based on each client’s data volume. Li et al.~\cite{42} introduced FedProx by adding a regularization term to the client loss function, aligning updates more closely with the global model. 
However, the one-global model-fits-all approach ignores client data diversity. To solve this, our approach extends FedProx by distributing a personalized global model to each client.


\section{Proposed Psychological Study on AU Incoordination}\label{sec:psyExp}
This study employed a behavioral experiment to investigate individuals’ cognitive responses to AU incoordination in the upper and lower facial regions. 
\par
The study included 30 college participants (20 female, \textit{Mean}= 22.22 years), a sample size considered adequate for statistical reliability under the Central Limit Theorem~\cite{kwak2017central}. All procedures followed the Declaration of Helsinki and were approved by the institutional review board of the Institute of Psychology, Chinese Academy of Sciences.
\par
\begin{figure}[t]
    \centering
    \begin{subfigure}{0.18\linewidth}
        \centering
         \includegraphics[width=0.9\linewidth, height = 1.8cm]{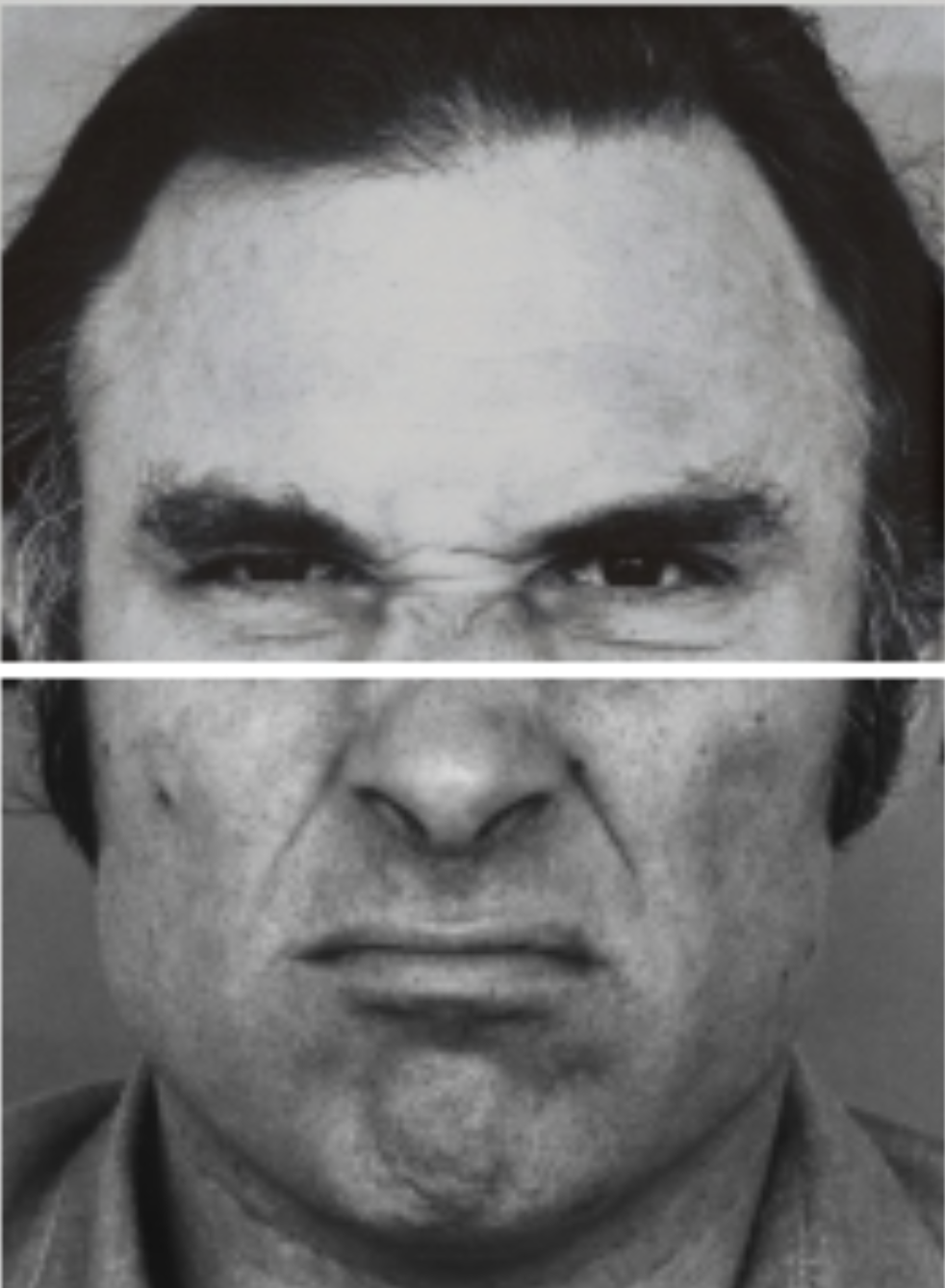} 
        \caption{AU 4+9}
    \end{subfigure}
 \hfill
    \begin{subfigure}{0.18\linewidth}
        \centering
          \includegraphics[width=0.9\linewidth,height = 1.8cm]{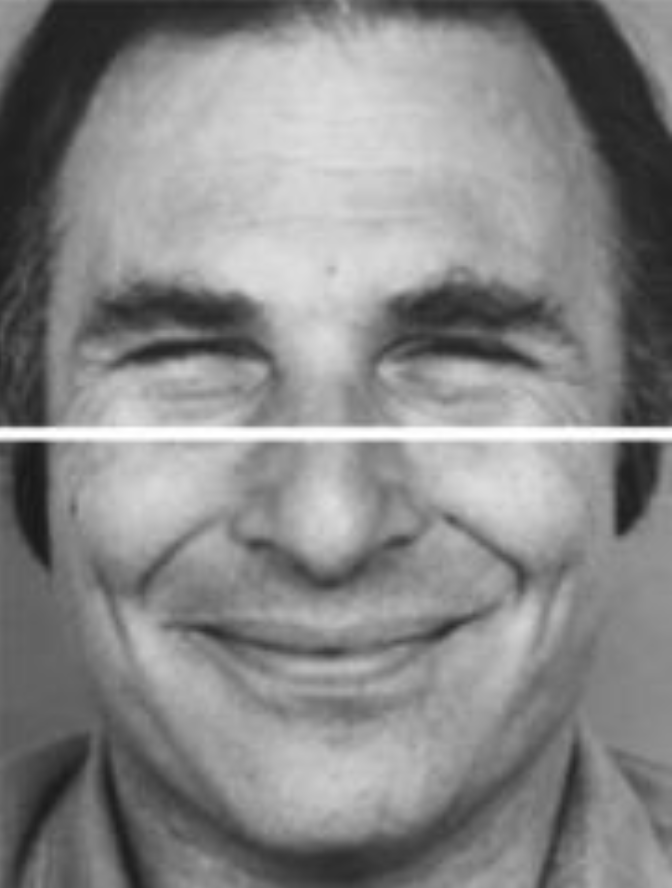} 
        \caption{AU 6+12}
    \end{subfigure}
 \hfill
     \begin{subfigure}{0.18\linewidth}
        \centering
         \includegraphics[width=0.9\linewidth,height = 1.8cm]{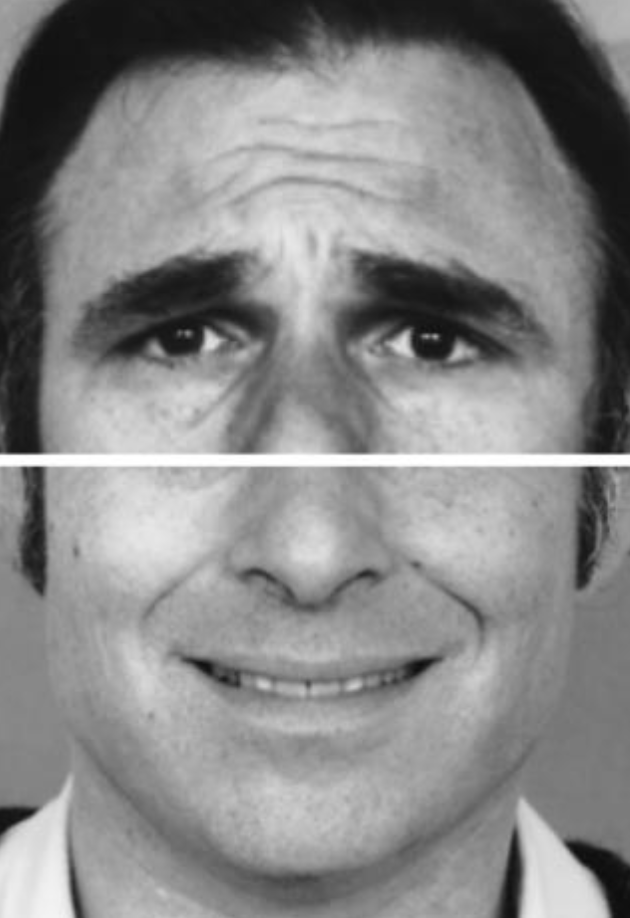} 
        \caption{AU 4+12}
    \end{subfigure}
\hfill
      \begin{subfigure}{0.18\linewidth}
        \centering
         \includegraphics[width=0.9\linewidth,height = 1.8cm]{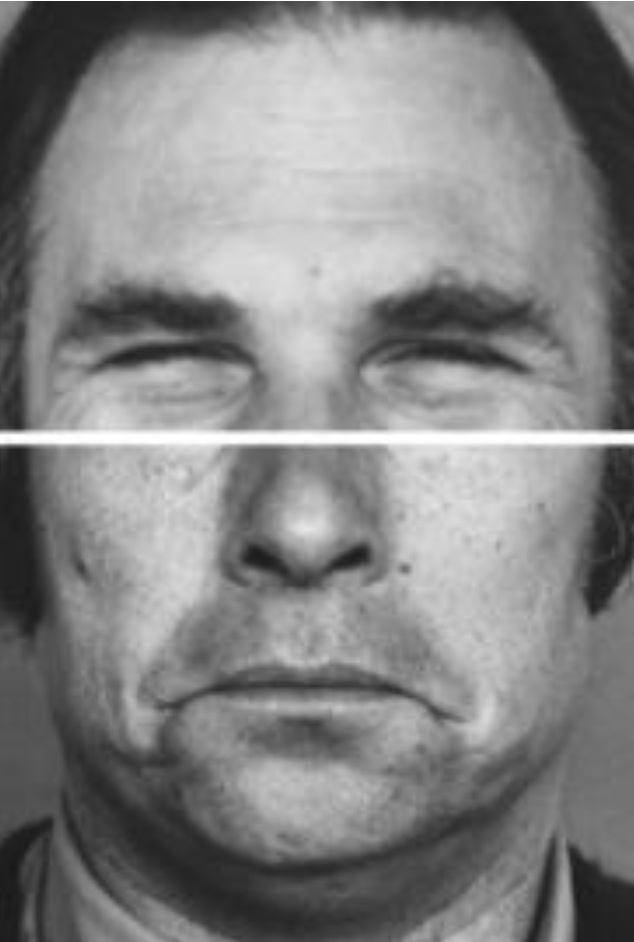} 
        \caption{AU 6+15}
    \end{subfigure}
 \caption{Examples of AU composite images. (a) and (b) are coordinated composites. (c) and (d) are uncoordinated composites.  }
    \label{fig:au_4exps}
\end{figure}

This behavioral experiment employed a within-subjects design. The independent variable was the type of AU composite image (coordinated vs. uncoordinated AU composites), as shown in~\cref{fig:au_4exps}. The face images were segmented anatomically. The dividing line was consistently present in all stimuli, serving as a control variable and would not influence the experiment's outcomes. To control for individual facial differences in emotion recognition, the original images used for composites were sourced from the same person~\cite{ekman1978facial}. 
Seventy-two composite images were generated from upper and lower facial AUs across six basic emotions, with each emotion represented by six coordinated and six uncoordinated expressions.

\par
The dependent variables are accuracy rates and scores. Specifically, participants made two judgments for each composite image: whether it was coordinated or uncoordinated, and the degree of coordination or incoordination.

\par
Statistical analysis from this study showed that participants were significantly more accurate in recognizing coordinated AU composite images than uncoordinated ones (Please see Supplementary Materials (Suppl.) for more experimental details and statistical results).
The study further highlighted specific AU combinations, such as AU4 and AU12, that participants quickly identified as uncoordinated, whereas combinations like AU6 and AU12 were rapidly recognized as coordinated. This suggests that these uncoordinated AU pairings may inherently convey emotional incoordination. Besides, the coordination patterns between specific AUs reveal systematic facial action structures. These patterns help the network distinguish both coordinated and uncoordinated facial movements, thus enhancing its ability to identify ME features more effectively.

\section{Our Proposed FED-PsyAU Framework}

\begin{figure*}[ht]
    \centering
    \includegraphics[width=\linewidth]{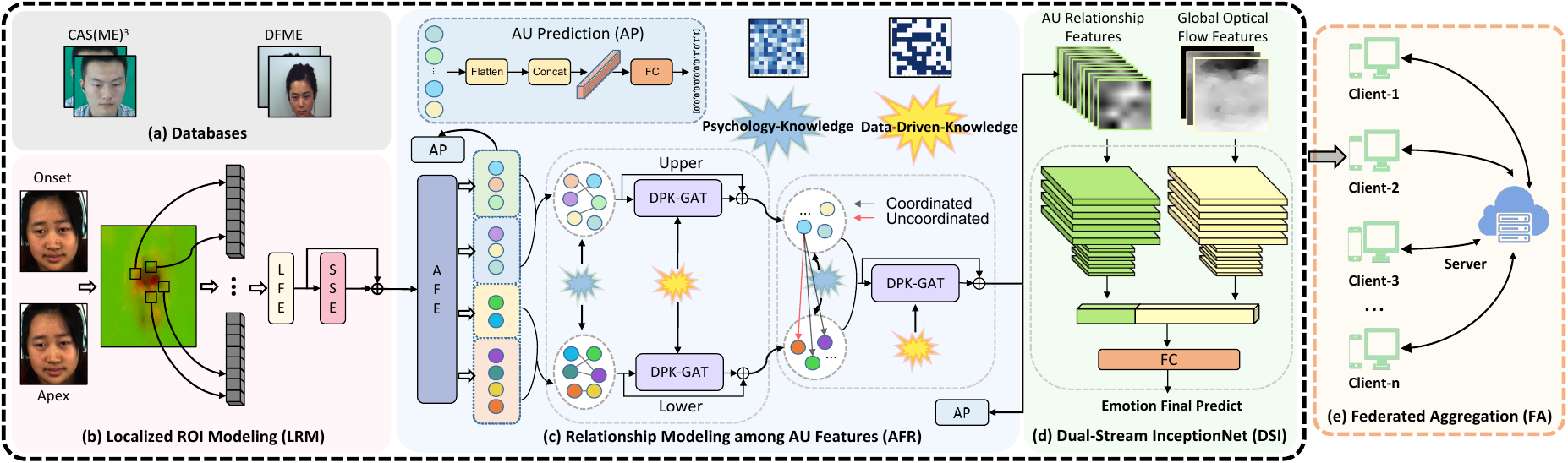}
    \caption{The pipeline of our proposed FED-PsyAU framework, including MER network (black block) and the Federated Aggregation (FA) module (orange block). In MER network, the LRM module consists of the local ROI feature extractor (LFE) and the spatial structure encoder (SSE), which preserves spatial structure information while efficiently extracting ROI features; the AFR module includes the AU feature extractor (AFE) and the dynamic prior knowledge-based graph attention network (DPK-GAT) for effective AU feature extraction and capturing AU dynamic relationships, the DSI module use a dual-stream structure to capture multi-scale facial muscle topology and motion information for MER. Then, the FA module integrates the MER network into the FL framework.}
    \label{fig:model_overview}
\end{figure*}

We propose the FED-PsyAU framework (\cref{fig:model_overview}), which embeds a psychology-driven MER model within a federated aggregation module. The model features a hierarchical structure that learns from local ROIs to AU groups and finally to global facial regions, capturing dynamic topology while minimizing redundancy. It leverages a multi-scale InceptionNet to extract and fuse global OF with structural facial features. The federated module enables privacy-preserving collaborative training to boost MER performance on each client.
\subsection{Localized ROI Modeling}

 

\begin{figure}[t]
    \centering
    \includegraphics[width=0.95\linewidth]{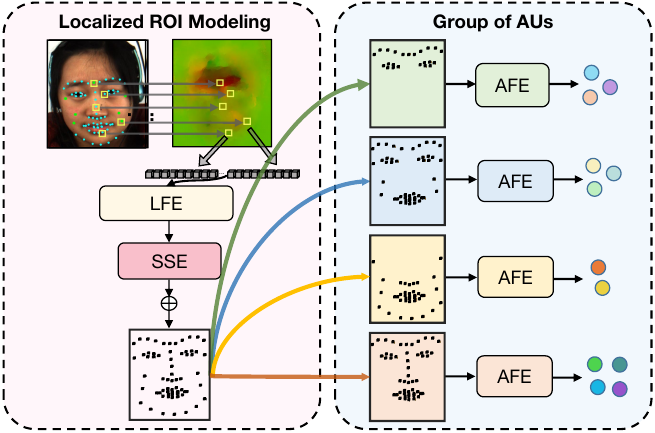}
    \caption{ROI grouping strategy for AU. OF features from each ROI are processed through LFE and SSE to yield high-dimensional representations. ROIs are then grouped by AU movement patterns to create inputs for AFEs, ultimately producing AU node features for the GAT network.}
    \label{fig:roi_au_feature_extract}
\end{figure}

Since MEs are manifested by muscle movements in key facial regions (e.g., eyes, mouth, nose)~\cite{ekman1978facial}, we refine the standard 68 landmarks from Dlib~\cite{87} to precisely capture these changes. As shown in \cref{fig:roi_au_feature_extract}, we discard 10 less informative outer contour landmarks and add 7 new midpoints to better track movements in the cheek and eyebrow regions, resulting in a final set of 65 keypoints.

\par
Then, onset-apex OF inputs ($128 \times 128$ pixels) are cropped into 65 ROIs, each measuring $5 \times 5$ pixels, centered on selected facial landmarks $\mathbf{p}_{k} = (x_{k}, y_{k})$, where $k \in {1, 2, \dots, 65}$, denotes the landmark index. To enhance robustness to apex annotation, we also compute OF using a small window of frames around the apex. The ROI features input to the \textbf{L}ocal ROI \textbf{F}eature \textbf{E}xtractor (LFE), denoted as $\Theta_{k}$, contain the horizontal and vertical components of OF and optical strain.
The LFE module consists of three convolutional layers, each followed by batch normalization and ReLU activation. All kernel sizes were $3 \times 3$. The output channels in the first two layers increase from 32 to 64, while the final layer outputs 3 channels. This design helps the LFE capture complex local patterns, with the output features referred to as $\Phi_{k} = \text{LFE}(\Theta_{k})$.


Inspired by Vision Transformer~\cite{59}, we use a \textbf{S}patial \textbf{S}tructure \textbf{E}ncoder (SSE), i.e., Transformer Encoder with three attention heads to capture diverse dependencies among ROIs, as MEs result from coordinated muscle movements across multiple areas. 
Precisely, we convert $\Phi_{k}$, into a token. To retain the positional information of the facial landmarks, a positional embedding $E_k$ is added based on the landmark index to each token, yielding $X_{k} = \Phi_{k} + E_k$.
We add the output of the three head attention module to the input feature $X_{k}$ , and then normalize the result using LayerNorm.
Finally, after passing through a feed-forward neural network and applying LayerNorm, the final output representation is obtained as $Z_k \in \mathbb{R}^{ 3 \times 5 \times 5}$.

\subsection{Relationship Modeling among AU Features}
Our proposed Relationship Modeling among \textbf{A}U \textbf{F}eatures (AFR) module consists of two components: an \textbf{A}U \textbf{F}eature \textbf{E}xtractor (AFE) to derive features from AU-specific facial regions, and a GAT-structure to model the topological relationships between these AUs.

\subsubsection{AU Feature Extractor}\label{subsec:AFE}
\par
MEs involve dynamic, coordinated interactions across multiple regions~\cite{ekman1978facial} and are characterized by subtle movements that differ from typical macro-expressions. Therefore, as shown in \cref{fig:roi_au_feature_extract}, when grouping ROIs for the AFE module, the AU selection and regionalization are  based on: 1) AU frequency in MEs; 2) Grouping by primary/secondary muscle activation areas~\cite{dong2022spontaneous}. Notably, not all AUs employed in Section~\ref{sec:psyExp} are applied to MER. We specifically retain only the 12 AUs exhibiting the highest relevance to ME dynamics, as listed in \cref{tab:au_selection}. Detailed occurrences are provided in the Suppl.


\begin{table}[t]
    \centering
    \caption{Selected AUs and Corresponding Regions (Primary and Secondary/ Coordinate). $g$, $N_g$ and $N_{g\_AU}$ represent ROI Group index, corresponding ROI number and AU amount.}
    \label{tab:au_selection}
    \resizebox{\linewidth}{!}{
    \begin{tabular}{cccccc}
        \toprule
        \textbf{\textit{g}} & \textbf{AU} & \textbf{Primary} & \textbf{Secondary/ Coordinate} & $\bm{N_g}$& $\bm{N_{g\_AU}}$\\
        \midrule
        1 & 1,2,4 & Eyebrows & Eyes & 23&3 \\
        2 & 5,6,7 & Eyes & Eyebrows, Cheeks, Mouth & 48& 3\\
        3 & 9,10 & Nose, Cheeks & Mouth, Chin & 38&2 \\
        4 & 12,14,15,17 & Mouth & Eyes, Eyebrows, Nose & 52&4 \\
        \bottomrule
    \end{tabular}}
\end{table}

\par


The first step we construct in AFE is \textbf{G}roup \textbf{S}queeze and \textbf{E}xcitation (GSE). It computes the attention for each group of AUs (ROI group), which makes the model pay more attention to the critical ROIs. Specifically, for each ROI group input $R_g = \{Z_1, Z_2, \dots, Z_{N_{g}}\}, \; g \in \{1, 2, 3, 4\}$, the GSE module performs global average pooling on each ROI and computes channel importance through a fully connected (FC) layer. The original features are then preserved through residual concatenation: 
\begin{equation}
    f_{\text{GSE}}(R_g) = \{ W^{att}_k \odot Z_k + Z_k \mid Z_k \in R_g \}
\end{equation}
where $ W^{att}_k $ is the attention weight for each group of ROI, $\odot$  denotes element-wise multiplication along the channel dimension of $Z_k$.
Next, $f_{\text{GSE}}(R_g)$ undergoes further feature extraction, reducing from ${N_g}$ composite features to $N_{g\_AU}$ representative AU features, yielding the output feature $F_{g\_AU}$.
Precisely, the process comprises two convolutional layers: a $3\times3$ layer that doubles the input dimensions and a $1\times1$ layer that restores them to the original size. Both layers are followed by Batch Normalization and ReLU activation for improved stability and representation.


\subsubsection{AU Relationship Modeling}\label{AUs Relationship Modeling}
\textbf{Prior Knowledge:}
%
To model AU dynamic relationships, we integrate two forms of prior knowledge: an ``intrinsic" prior from psychological principles and an ``extrinsic"  prior from data-driven patterns. The intrinsic prior, derived from our psychological findings on AU coordination (\cref{sec:psyExp}), provides structural guidance by defining the initial GAT adjacency matrix $A$. The extrinsic prior, based on statistical AU co-occurrence patterns in the data, offers data-driven refinement that theoretical models may overlook. This extrinsic prior forms a prior attention matrix $D$ to guide weight allocation among AU nodes. By fusing this foundational psychological structure with real-world dynamics, our model achieves superior adaptability and generalization (see Suppl. for matrix details).

\noindent\textbf{Facial Segmentation:}
As presented in~\cref{sec:psyExp}, the upper and lower facial regions, controlled by different neural pathways, exhibit distinct muscle movement patterns. AUs and their associated primary muscles in the upper and lower face can exhibit coordination (potential co-occurrence) or lack of coordination (infrequent simultaneous appearance). Leveraging these intra- and inter-regional interactions, we model AU relationships for the upper face, lower face, and whole face. Consistent with the experimental setup in~\cref{sec:psyExp}, we segment the face into upper and lower parts based on the orbicularis oculi muscle.
Specifically, following the primary occurrence regions of 12 listed AUs, the AUs in the first two groups from \cref{tab:au_selection} are classified as upper face AUs, while those in the last two groups are classified as lower face AUs.
\begin{equation}
    F_{AU}^{Upper} = F_{1\_AU} \cup F_{2\_AU} \text{,~} F_{AU}^{Lower} = F_{3\_AU} \cup F_{4\_AU} 
\end{equation}
\par
We input the AU features from the upper and lower face: $F_{AU}^{Upper}$, $F_{AU}^{Lower}$ into our proposed \textbf{D}ynamic \textbf{P}rior \textbf{K}nowledge GATs (DPK-GATs) in parallel (Details are presented in the next part of this subsection). These networks use psychology prior knowledge to build AU adjacency relationships and data-driven priors to guide attention learning, producing outputs $h^{Upper}$ and $h^{Lower}$, which are combined into $h^{Local}$ for the next stage.
\par
\begin{figure}[t]
    \centering
    \includegraphics[width=\linewidth]{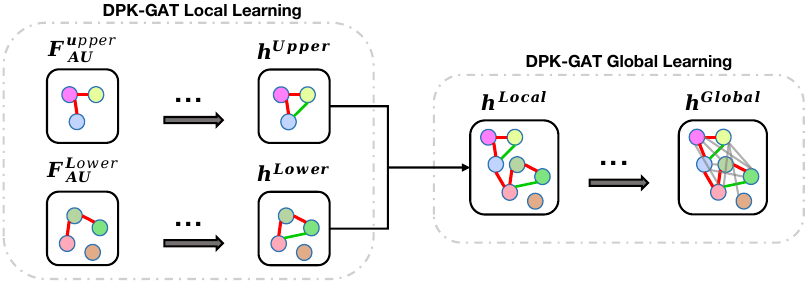}
    \caption{ DPK-GAT from facial segmentations to global face.
    First, intra-regional AU node relationships are learned within the upper and lower facial regions. Second, cross-regional relationships are modeled to integrate interactions between these regions.
    }
    \label{fig:local_global_GAT}
\end{figure}

In the whole-face DPK-GAT, we construct the network similarly but set adjacency values within the same region to zero, ignoring intra-regional AU relationships. By learning cross-regional topological structures and statistical patterns, we obtain global AU relationship features, denoted as $h^{Global}$, as illustrated in \cref{fig:local_global_GAT}.

\par
\noindent\textbf{DPK-GAT:}
We model dynamic AU node relationships combing GAT and 
the prior knowledge. In particular, for the input feature belonging to $\{ F_{AU}^{Upper}, F_{AU}^{Lower}, h^{Local} \}$, a linear mapping projects it to a higher-dimensional space as $f$. The similarity score \( e_{ij} \) between AU nodes \( i \) and \( j \), is calculated using the attention weight vector \( \vec{a} \) and concatenated node features \( f_i \) and \( f_j \):
\begin{equation}
e_{ij} = \begin{cases}
\text{LeakyReLU} \left( \vec{a}^T [f_i || f_j] \right), & \text{if } A_{ij} > 0 \\
-\infty, & \text{if } A_{ij} = 0
\end{cases}
\end{equation}
Here, only neighboring nodes defined by the psychology-derived adjacency matrix \( A \) are considered.

The final attention $\alpha_{ij}$ is calculated by combining the data-driven prior attention  $D_{ij}$ and the self-attention, obtained by the softmax operation applied to $e_{ij}$:
\begin{equation}
\alpha_{ij} = (1 - \beta) \cdot \text{softmax}(e_{ij}) + \beta \cdot D_{ij}
\end{equation}
where, $\beta$ is the dynamic prior attention weight, updated based on the training rounds.
Along with a residual concatenation, the final node representation $h_i$ is aggregated as:
\begin{equation}
h_i = \text{ELU} ( \textstyle \sum_{j \in \mathcal{N}(i)} \alpha_{ij} \cdot f_j ) + f_i
\end{equation}
where $\mathcal{N}(i)$ denotes the neighboring nodes of $i$. In our implementation, we employ three attention heads and a two-layer structure, enabling the extraction of complex AU dependencies.

\subsection{Dual-Stream-InceptionNet}

Muscle movements vary widely across facial regions, making multi-scale information essential for the MER task. InceptionNet’s core concept is leveraging parallel convolutional kernels to extract multi-scale features, enhancing the network’s ability to capture fine details and overall structure~\cite{63}. We build on this by using a dual-stream InceptionNet as the ME feature extractor. One stream processes the AU relationship features $h^{Global}$, while the other handles global OF features $\Theta$. This dual-stream design allows the model to capture the topological relationships of facial muscles through AUs and holistic motion via OF. The output tensors from both streams are flattened, concatenated, and fed into a FC layer for ME prediction.

\subsection{Federated Aggregation}

Traditional centralized learning consolidates all data on a single server, enhancing model accuracy and generalization but neglecting privacy concerns in MER. FL could address this by training models locally and sharing only parameters, protecting data privacy and retaining each client’s unique data characteristics.
\par
FedProx~\cite{64} is a FL approach designed to address data heterogeneity among clients. It incorporates proximal terms during model aggregation to balance global and client-local learning, improving both performance and convergence speed. We extend FedProx by altering the server's processing and distribution strategy during aggregation, which we call \textbf{P}ersonalized \textbf{FedProx} (P-FedProx). Specifically, P-FedProx initializes each client’s model $W_i^{t}$ in round $t$ by combining its previous model $\mathring{W}_i^{t-1}$ with models from other clients $\mathring{W}_j^{t-1}$.
\small
\begin{equation}
W_i^t = \theta \cdot \mathring{W}_i^{t-1} + \textstyle \sum_{\substack{j=1 \\ j \neq i}}^{n} \omega_j \cdot \mathring{W}_j^{t-1} \text{,~where} \sum_{\substack{j=1 \\ j \neq i}}^{n} \omega_j = 1 - \theta
\end{equation}
\normalsize
where weight $\omega_j$ is proportional to $j$th client's data size. For $i$th client, $\theta$ is set to 0.9 (optimal experimental result, see more detail in the Suppl.)

\subsection{Loss Function}
\textbf{AU Prediction Loss:}
To enhance MER, our network models the topological relationships and statistical patterns of AUs, making AU prediction an auxiliary task. The prediction is utilized in two stages: one enhances the AU feature extraction performance of the AFE module, and the other improves the GAT network's ability to learn AU topological structures. AU recognition is formulated as a multi-label classification problem, with the Binary Cross-Entropy Loss for each AU label defined as:
\small
\begin{equation}
L_{\text{A}} = \frac{\textstyle\sum_{j=1}^{M} \left[y_j  \cdot \log(\sigma(\hat{y}_j)) + (1 - y_j) \cdot \log(1 - \sigma(\hat{y}_j))\right]}{-M}
\end{equation}
\normalsize
where $ M $ is the total number of AUs, $ y_j $ , $ \sigma(\hat{y}_j) $ and $ \hat{y}_j $ represent respectively the ground truth label (0 or 1), the logit output and prediction after applying the sigmoid function for the $j $-th AU. The loss function is consistent for both stages, i.e., $ L_{\text{A}}^{\text{AFE}} $ and $ L_{\text{A}}^{\text{GAT}} $.
\par
\noindent\textbf{MER Loss:}
The Cross-Entropy Loss function is utilized for emotion classification.
Thus, our overall MER loss is composed as follows:
\small
\begin{equation} 
L_{\text{MER}} = \alpha_1 \cdot ( -\textstyle\sum_{i=1}^{N} y_i \cdot \log(\hat{y}_i) ) + \alpha_2 \cdot L_{\text{A}}^{\text{AFE}} + \alpha_3 \cdot L_{\text{A}}^{\text{GAT}} 
\end{equation}
\normalsize
where,  $N$ represents the number of classes, $y_i$ and $\hat{y}_i$ are the one-hot encoded ground truth label and the predicted probability for class $i$, $\alpha_1$, $\alpha_2$ and $\alpha_3$ are hyperparameters, set to 0.2, 0.8, and 0.8 to prioritize MER loss while balancing AU recognition (See parameter analysis in Suppl.). 
\par
\noindent\textbf{Federated Loss:}
During federated training, the local client’s loss function includes the MER loss \( L_{\text{MER}} \) and a proximal term:  
\begin{equation}
L_{\text{FL}} = L_{\text{MER}} + \frac{\alpha_4}{2}\cdot||W - W_t||^2
\end{equation}
where $W_t$ denotes the global model in the server's $t$-th round of distribution, and $W$ represents the current model of the client. $\alpha_4$ is the hyperparameter for the weight of the proximal term.
\section{Experiments}
\subsection{Configuration}\label{sec:conf}
\textbf{Databases:} We use two recently published large scale and challenging ME databases: CAS(ME)$^3$ and DFME. CAS(ME)$^3$-Part A offers 860 samples from 100 participants. DFME includes three subsets with (participants, samples): A (72,1,118), B (92,969), C (492,5,439). For Apex-first-frame scenario, we compute the OF between the apex and mid-frame towards the offset.

\noindent\textbf{Pre-training:} To enhance the model's ability to perceive facial motion patterns, we first pre-trained the AU module (LRM+AFR) on the DISFA~\cite{mavadati2013disfa} and CK~\cite{Kanade2000ck} datasets.
The effect of this pre-training is detailed in the Suppl.

\noindent\textbf{MER Experiment Settings:}
The evaluation follows standard protocols on two datasets. On CAS(ME)$^3$, we perform 3-class classification (positive, negative, surprised)~\cite{23} with Leave-One-Subject-Out (LOSO) validation.  On DFME, we conduct 7-class classification (Happiness, Surprise, Disgust, Sadness, Anger, Fear, and Contempt) using scale-adapted 10-fold cross-validation~\cite{18}. To address class imbalance, performance is measured by the UF1 and UAR. Experiments were implemented in PyTorch 1.13.0 and run on five NVIDIA GeForce RTX 4090 GPUs.
\par
\noindent\textbf{Federated Experiment Settings:}
The DFME and CAS(ME)$^3$ datasets are randomly split into 5 and 2 local clients, respectively, with each client assigned an equal number of subjects. However, variations in micro-expression (ME) counts per subject result in differing ME data distributions and quantities across clients, mirroring real-world data heterogeneity (See detailed data distributions in the Suppl.). Traditional LOSO and k-fold cross-validation can leak training samples in a federated setting, so we employ a random partition strategy with 70\% of samples for training and 30\% for testing, averaging results over 10 tests to minimize randomness. 
Furthermore, we assume an ideal FL scenario with no client dropouts and equal local updates per client. In FedAvg~\cite{41} and FedProx~\cite{42}, the server assigns an identical global model to all clients each round; differently, FedProx adds a proximal term to reduce model bias in local training. Our P-FedProx enhances this by assigning tailored initial models, with 90\% ($\theta$) weight on each client’s own model and 10\% on others'. Additionally, we compare against recent personalized FL methods: FedRep~\cite{collins2021exploiting} uses shared representations with personalized heads, ELLP~\cite{zhou2024efficient} employs parameter decoupling with Fisher Information weighting, and FedAS~\cite{yang2024fedas} applies parameter alignment and client synchronization. Unlike these approaches requiring architectural changes or complex procedures, P-FedProx achieves personalization through simple weighted initialization.

\subsection{Comparison to State-of-the-art Methods}
\cref{tab:dfme_cas3} list the SOTA comparisons on DFME and CAS(ME)$^3$. The confusion matrices are shown in the Suppl.
\par
Regarding DFME, classifying ME samples into seven categories poses significant challenges, resulting in relatively low UF1 and UAR scores across all methods. However, our proposed method outperforms all others in every metric. Specifically, Our method outperforms I3D by 9.30\% in UF1 and 9.20\% in UAR, and surpasses LBP-TOP by 15.17\% in UF1 and 13.25\% in UAR, highlighting its effectiveness in extracting critical and discriminative features from ME clips. Additionally, compared to recent deep learning methods like FR, our approach yields 2.94\% and 1.64\% higher UF1 and UAR, confirming the benefits of incorporating facial topology and structured muscle motion for enhanced performance.
\par
Regarding CAS(ME)$^3$, the primary challenge lies in its sample complexity. Compared to baseline methods (STSTNet, RCN-A, FR), our proposed method achieves over approximately 20\% higher UF1 and UAR scores, marking a significant advancement. Our method surpasses three recent SOTA approaches in MER, including HTNet, which employs a hierarchical Transformer for local feature learning via self-attention. Specifically, our approach outperforms HTNet by 4.54\% in UF1 and 8.11\% in UAR, leveraging psychological insights and statistical patterns to capture deeper facial muscle dynamics. Additionally, it exceeds Micro-BERT and HSTA by 6.17\% and 2.91\% in UF1, and 1.01\% and 0.46\% in UAR, respectively. By mining facial movement patterns to better understand topology, our model achieves improved metrics, indicating enhanced accuracy in classifying all emotion categories .

\begin{figure*}[htbp]
    \centering
 \includegraphics[width=.45\linewidth]{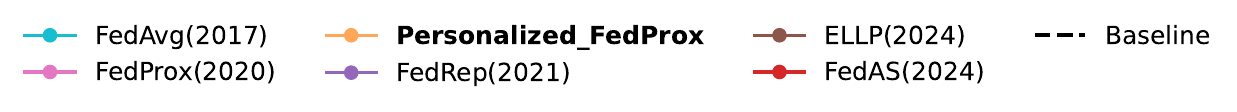}
\end{figure*}
\begin{figure*}[htbp]
    \centering
    \begin{subfigure}{\textwidth}
        \centering
        \includegraphics[width=0.95\textwidth]{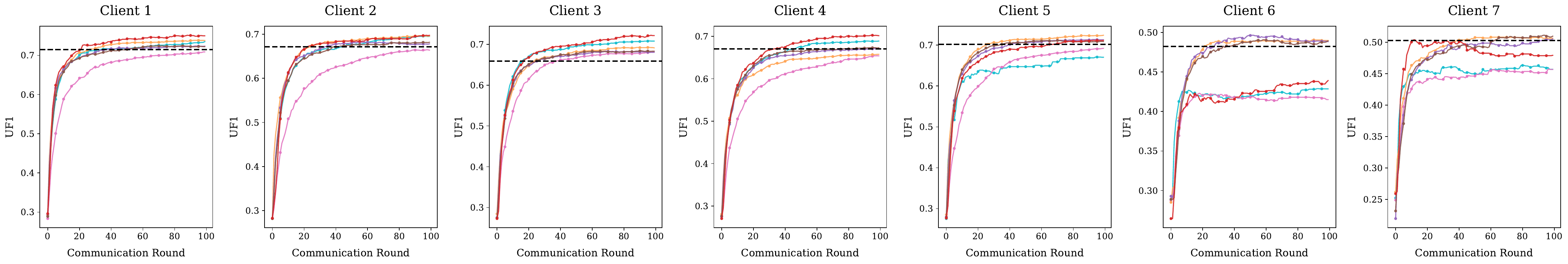}
        \caption{UF1 comparison across different clients.}
        \label{fig:uf1}
    \end{subfigure}

    \begin{subfigure}{\textwidth}
        \centering
        \includegraphics[width=0.95\textwidth]{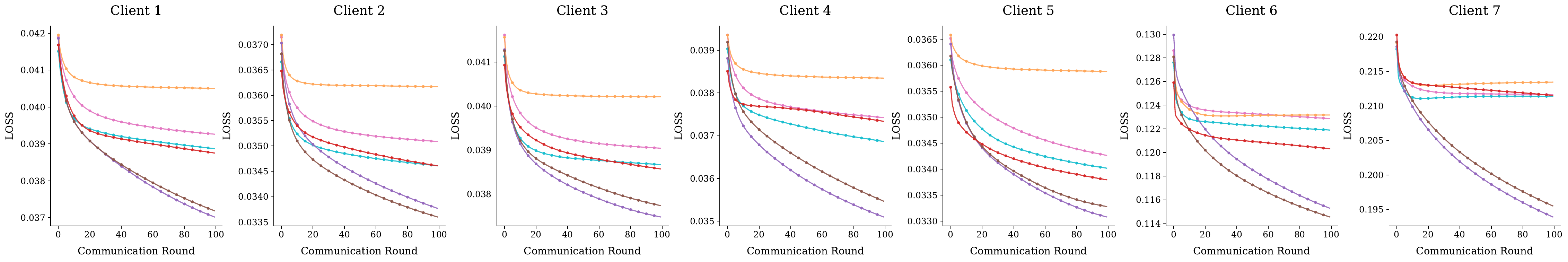}
        \caption{Convergence comparison across different clients.}
        \label{fig:loss}
    \end{subfigure}
    \caption{Compare the performance of different FL frameworks on different clients: UF1 and convergence. See UAR comparison in Suppl.}
    \label{fig:fedExp}
\end{figure*}

\subsection{Ablation Study}

\cref{tab:Ablation_module} listed the ablation study results on DFME.
\begin{table}[]
    \caption{SOTA methods comparison on DFME and CAS(ME)$^3$}
    \label{tab:dfme_cas3}
    \centering
    \resizebox{\linewidth}{!}{
    \begin{tabular}{c l ccccc}
        \toprule
        \multirow{2}{*}{ \textbf{Category}}& 
        \multirow{2}{*}{\textbf{Method}} & \multirow{2}{*}{\textbf{Year}} & \multicolumn{2}{c}{\textbf{DFME}} & \multicolumn{2}{c}{\textbf{CAS(ME)$^3$}} \\
         \cmidrule(lr){4-5} \cmidrule(lr){6-7}
         & & & \textbf{UF1} & \textbf{UAR} & \textbf{UF1} & \textbf{UAR} \\
        \midrule
        \multirow{2}{*}{\parbox{2cm}{\centering \textbf{\textit{3D-CNN}}}} & I3D~\cite{76} & 2017 & 0.2923 & 0.3058 & -- & -- \\
        & R3D~\cite{75} & 2018 & 0.2164 & 0.2313 & -- & -- \\
        \midrule
        \multirow{2}{*}{\parbox{1.5cm}{\centering \textbf{\textit{Hand-crafted}}}}& LBP-TOP~\cite{LBP-TOP} & 2014 & 0.2336 & 0.2653 & -- & -- \\
        & MDMO~\cite{77} & 2015 & 0.2489 & 0.2939 & -- & -- \\
        \midrule
        \multirow{13}{*}{\parbox{2cm}{\centering \textbf{\textit{Deep Learning}}}} & OFF-ApexNet~\cite{78} & 2019 & 0.2386 & 0.2806 & -- & -- \\
        & AlexNet~\cite{83} & 2012 & -- & -- & 0.2570 & 0.2634 \\
        & STSTNet~\cite{25} & 2019 & 0.2714 & 0.3108 & 0.3795 & 0.3792 \\
        & RCN-A~\cite{79} & 2020 & 0.2751 & 0.3123 & 0.3928 & 0.3893 \\
        & MERSiam~\cite{80} & 2021 & 0.3184 & 0.3532 & 0.3184 & 0.3532 \\
        & FGRL~\cite{34} & 2021 & 0.0736 & 0.1429 & 0.3333 & 0.2636 \\
        & FR~\cite{34, 28} & 2022 & \underline{0.3559} & \underline{0.3814} & 0.3493 & 0.3413 \\
        & MMNet~\cite{81} & 2022 & 0.2649 & 0.2776 & 0.3706 & 0.3646 \\
        & BDCNN~\cite{67} & 2022 & 0.2975 & 0.3314 & 0.5050 & 0.5164 \\
        & Micro-BERT~\cite{84} & 2023 & -- & -- & 0.5604 & 0.6125 \\
        & HTNet~\cite{71} & 2024 & 0.3243 & 0.3368 & 0.5767 & 0.5415 \\
        & HSTA~\cite{85} & 2024 & -- & -- & \underline{0.593} & \underline{0.618} \\
        & \textbf{Ours} & - & \textbf{0.3853} & \textbf{0.3978} & \textbf{0.6221} & \textbf{0.6226} \\
        \bottomrule
    \end{tabular}}
\end{table}

\begin{table}
    \caption{Ablation study. L, G, S and D represent Local (upper and lower face), Global, Single stream and Dual Stream, respectively. } 
    \label{tab:Ablation_module}
    \centering
    \resizebox{\linewidth}{!}{
    \begin{tabular}{@{\extracolsep{\fill}}c c c c c c c c c} 
        \toprule
        \multirow{2}{*}{\centering\textbf{\quad Setting}} & \multicolumn{2}{c}{\textbf{LRM}} & \multicolumn{2}{c}{\textbf{AFR}} & \multicolumn{2}{c}{\textbf{InceptionNet}} & \multirow{2}{*}{\textbf{UF1} $\uparrow$} & \multirow{2}{*}{\textbf{UAR} $\uparrow$} \\
        \cmidrule(lr){2-3} \cmidrule(lr){4-5} \cmidrule(lr){6-7}
        & \textbf{LFE} & \textbf{SSE} & \textbf{L} & \textbf{G} & \textbf{S} & \textbf{D} & & \\
        \midrule
        \multicolumn{1}{c}{\textrm{I}}   &  &  &  &  &\checkmark  &  & 0.3249 & 0.3404 \\
        \multicolumn{1}{c}{\textrm{II}}  &\checkmark  &  &  &  &  &\checkmark &0.3362 & 0.3517 \\
        \multicolumn{1}{c}{\textrm{III}}  &\checkmark  &\checkmark  &  &  &  &\checkmark &0.3480 & 0.3624 \\
        \multicolumn{1}{c}{\textrm{IV}}   &\checkmark  &\checkmark  &\checkmark  &  &  &\checkmark &0.3629 & 0.3723 \\
        \multicolumn{1}{c}{\textrm{V}}   &\checkmark  &\checkmark  &  &\checkmark  &  &\checkmark &0.3604 & 0.3658 \\
        \multicolumn{1}{c}{\textrm{VI}} &\checkmark  &\checkmark  &\checkmark  &\checkmark  &  &\checkmark &\textbf{0.3853} & \textbf{0.3978} \\
         \bottomrule
         \rule{0pt}{12pt}\textbf{Setting}&  \multicolumn{3}{c}{\textbf{Without AU Group}}& \multicolumn{3}{c}{\textbf{ With AU Group}}& \textbf{UF1} $\uparrow$ & \textbf{UAR} $\uparrow$ \\
       \midrule
        \textrm{VII}   & \multicolumn{3}{c}{\checkmark} &\multicolumn{3}{c}{}  & 0.3683 & 0.3793 \\
        \textrm{VIII}  & \multicolumn{3}{c}{}& \multicolumn{3}{c}{\checkmark} & \textbf{0.3853} & \textbf{0.3978} \\
       
        \bottomrule
       \rule{0pt}{12pt}\textbf{Setting}&  \multicolumn{3}{c}{\textbf{Psychological prior}}& \multicolumn{3}{c}{\textbf{ Data-driven Prior}}& \textbf{UF1} $\uparrow$ & \textbf{UAR} $\uparrow$ \\
       \midrule
        \textrm{IX}   & \multicolumn{3}{c}{} &\multicolumn{3}{c}{}  & 0.3615 & 0.3701 \\
        \textrm{X}  & \multicolumn{3}{c}{\checkmark}& \multicolumn{3}{c}{} & 0.3723 & 0.3832 \\
        \textrm{XI}  &\multicolumn{3}{c}{} &\multicolumn{3}{c}{ \checkmark}& 0.3705 & 0.3767 \\
        \textrm{XII}   & \multicolumn{3}{c}{\checkmark} & \multicolumn{3}{c}{\checkmark  }&\textbf{0.3853} & \textbf{0.3978} \\
        \bottomrule

    \end{tabular}
    }
\end{table}

\par
\noindent\textbf{Impact of AU Feature:} 
Setting \textrm{I}, which uses only full face OF as input, serves as our baseline for evaluating the role of AU features in MER. In Setting \textrm{II}, integrating AU features into the Dual Stream InceptionNet leads to a 1.13\% improvement in both UF1 and UAR. This improvement demonstrates that incorporating AU features significantly enhances the model’s ability to recognize MEs, and underscores the value of fusing multiple feature types to strengthen model performance.
\par
\noindent\textbf{Impact of ROI Relationship Modeling:} 
The results in Setting \textrm{III} show improvements of 1.18\% in UF1 and 1.07\% in UAR compared to Setting \textrm{II}, indicating that the model effectively captures complex dependencies between ROIs and extracts more discriminative feature representations.
\par
\noindent\textbf{Impact of AU grouping:} 
We compared AU feature extraction from the global face with our AU 
 grouping strategy (Setting \textrm{VII} vs. \textrm{VIII}), the performance improvement (UF1: 1.70\%, UAR: 1.85\%), confirming that AU grouping enables the network to learn more representative AU features. 
\par
\noindent\textbf{Impact of Modeling the Topological Structure of ME Muscle Movements:} 
Compared to Setting \textrm{III}, Setting \textrm{IV}, including local muscle movement topology modeling, improves UF1 and UAR by 1.49\% and 0.99\%, respectively. Setting \textrm{V}, which adds global muscle movement topology modeling, shows gains of 1.24\% in UF1 and 0.34\% in UAR over Setting \textrm{III}. Combining both local and global topology modeling yields the best performance on DFME. These results demonstrate that incorporating topological modeling of muscle movements enhances MER.
\par
\noindent\textbf{Impact of Prior Knowledge:}
\cref{tab:Ablation_module} presents the ablation experiments to evaluate the impact of using prior knowledge for constructing facial muscle topology. In particular, compared to Setting \textrm{IX}, Setting \textrm{X}, which uses only psychological knowledge, improves UF1 and UAR by 1.08\% and 1.31\%, respectively. Setting \textrm{XI}, incorporating only data-driven knowledge, shows gains of 0.9\% in UF1 and 0.66\% in UAR. When both psychological and data-driven knowledge are combined (Setting \textrm{XII}), the improvements reach 2.38\% in UF1 and 2.77\% in UAR, proving the effectiveness of the prior knowledge. Constructing a prior adjacency matrix allows the model to capture complex facial muscle topology, while data-driven knowledge directs attention to uncover intrinsic movements patterns of facial muscles.

\subsection{Federated Experiment}
Since FL aims to enhance MER performance on individual clients in scenarios with limited sample sizes and data privacy issue, and given that DFME and CAS(ME)$^3$ have large sample size, we partition the datasets as mentioned in \Cref{sec:conf} and treat the subdivisions as separate clients.
\par
While our focus is on application rather than fundamental FL innovation, our experiments confirm the effectiveness of our P-FedProx method, which is tailored for ME data distributions. Benchmarked against several open-source SOTA methods, our approach excels at mitigating small-sample challenges while preserving data privacy.
Specifically (\cref{fig:fedExp}),
the results show P-FedProx outperforms both the single-client baseline and traditional FL methods (FedAvg, FedProx), with particularly significant gains on clients with more challenging data distributions (Clients 6-7). Compared to recent personalized FL methods, our approach surpasses FedRep and ELLP and achieves performance competitive with the state-of-the-art FedAS, yet with a significantly simpler implementation. These consistent gains demonstrate that our weighted model initialization effectively addresses the data heterogeneity that challenges traditional FL, thereby mitigating ME sample scarcity while ensuring data privacy.

Meantime, we note that all FL methods achieved good performance on the local clients from DFME, which may be due to the fact that these clients have a similar amount of data and slight differences in data distribution, presenting fewer challenges for federated learning.
\par
Besides, we analyze the training loss progression of different federated algorithms over communication rounds. We found that P-FedProx demonstrates the smoothest convergence curves with minimal oscillations, particularly evident in Clients 1-5. On the other hand, P-FedProx converges significantly faster than other FL methods. This rapid early convergence suggests that personalized global model assignment provides clients with better initialization states.

\section{Conclusion}
MER is challenged by limited sample sizes, subtle features, and privacy concerns in real-world applications.
To address these issues, we conduct a psychological study to provide structured insights into facial muscle dynamics. We then develop a FED-PsyAU framework that integrates these findings with data-driven patterns for hierarchical learning of dynamic features, enhancing MER performance. Additionally, the FL framework improves MER across clients while preserving data privacy through iterative model updates. The comprehensive experiments validate our approach’s effectiveness.
In the future, we will integrate additional AUs to build a more comprehensive ME topology information and further advance MER in real-world applications through FL and semi-supervised approaches.

\section{Acknowledgment}
This research was partially funded by 1) the National Natural Science Foundation of China (62476269, 62276252); 2) the Youth Innovation Promotion Association CAS. 


{
    \small
    \bibliographystyle{ieeenat_fullname}
    \bibliography{main}
}

\end{document}